# Integrating BIM and UAV-Based Photogrammetry for Automated 3D Structure Model Segmentation

SIQI CHEN and SHANYUE GUAN

## ABSTRACT

The advancement of UAV technology has enabled efficient, non-contact structural health monitoring. Combined with photogrammetry, UAVs can capture high-resolution scans and reconstruct detailed 3D models of infrastructure. However, a key challenge remains in segmenting specific structural components from these models—a process traditionally reliant on time-consuming and error-prone manual labeling. To address this issue, we propose a machine learning-based framework for automated segmentation of 3D point clouds. Our approach uses the complementary strengths of real-world UAV-scanned point clouds and synthetic data generated from Building Information Modeling (BIM) to overcome the limitations associated with manual labeling. Validation on a railroad track dataset demonstrated high accuracy in identifying and segmenting major components such as rails and crossties. Moreover, by using smaller-scale datasets supplemented with BIM data, the framework significantly reduced training time while maintaining reasonable segmentation accuracy. This automated approach improves the precision and efficiency of 3D infrastructure model segmentation and advances the integration of UAV and BIM technologies in structural health monitoring and infrastructure management.

## INTRODUTION

Rail infrastructure is the backbone of modern society. With nearly 140,000 route miles in operation, the U.S. freight rail system is one of the largest and most efficient railroad network in the world [1]. This widespread transportation network makes extensive contributions to the U.S. economy. Maintaining railroads in good condition is essential to provide sustainable economic development. Damage or wear in components like rails and crossties can result in serious accidents, including derailments [2]. When freight trains are involved, these accidents may also cause hazardous material spills and environmental harm [3]. Thus, inspecting and monitoring rail infrastructure conditions is vital, and more research in this area is urgently needed.

---

Siqi Chen, Graduate Research Assistant, Email: chen4469@purdue.edu; Shanyue Guan, Ph.D., P.E., Assistant Professor, Email: guansy@purdue.edu; School of Construction Management Technology, Purdue University, West Lafayette, IN, 47907, U.S.A.

Traditional manual inspection of railways is time-consuming, labor-intensive, and often puts inspectors' safety at risk. Workers must enter the track area, which increases the chance of accidents, especially collisions with passing trains [4]. In contrast, using unmanned aerial systems (UASs) for inspection offers a safer and more efficient alternative [5]. Drones can capture high-resolution imagery to generate 3D reconstruction models without the need for on-track access, greatly reducing the interruptions to maintenance or regular train operations. Moreover, the data collected from the UAS can be processed automatically, allowing for faster and more consistent inspections compared to manual methods. To evaluate individual components conditions more accurately, segmenting the collected point cloud data based on key structural components of the track is a critical step [6]. Deep leaning neural networks offers a promising approach for automating this segmentation process, but it typically requires a large number of labeled training data. However, manual annotation is both time-consuming and labor-intensive [7]. By integrating synthetic point cloud models generated from BIM to the training process, we can ease the high demands for manual labeling [8]. This approach helps expedite the workflow and improves the overall process efficiency.

This paper presents a practical framework that combines unmanned aerial systems (UAS) with high-resolution 3D reconstruction for railroad condition assessment. A key innovation is the integration of synthetic BIM data with a small training set to enhance point cloud segmentation efficiency. The pipeline includes UAS-based data collection, 3D model generation, and semantic preprocessing. Introducing BIM-generated samples reduces manual labeling needs, shortens training time, lowers memory usage, and improves segmentation performance. The method is validated on a full-scale railroad scan in Indiana, with results showing accurate identification of major components using limited training data. The paper concludes with perspectives on advancing UAS-based 3D modeling for rail infrastructure.

## METHODOLOGY

We developed an efficient method for 3D segmenting railroad track by key structural components with the aid of synthetic BIM data. The framework of this method illustrated in Figure 1 mainly consists of the data collection and preprocessing, the neural network architecture employed for segmentation, the training procedures, and the evaluation of the model's performance steps. Each step is introduced in detail in the later sections of this paper.

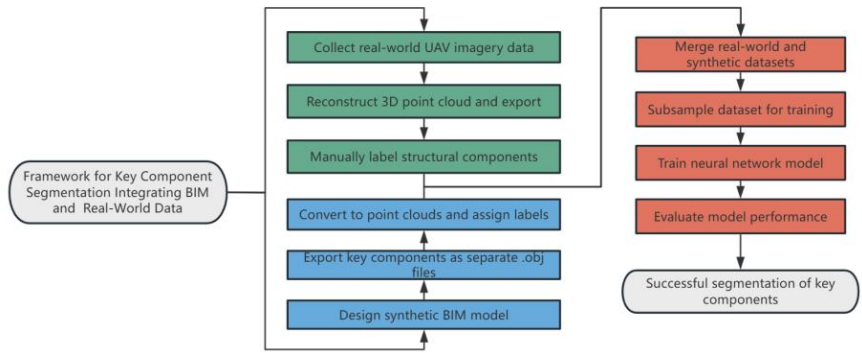

Figure 1. Framework for the segmentation of the railroad component

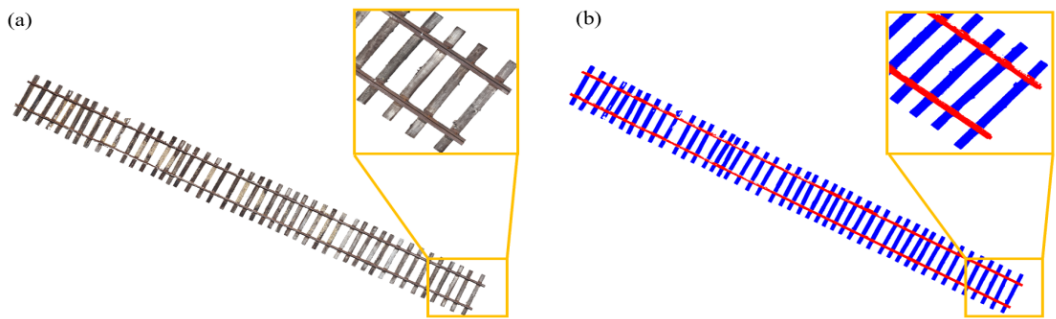

Figure 2. (a): 3D reconstruction of the railroad track with zoomed-in detail; (b): manually labeled point cloud highlighting rails (red) and crossties (blue) with zoomed-in detail.

**UAS-Based Real-World Railroad Data Collection and Manual Labeling**

A DJI Mavic 3 Thermal UAS was used to monitor the condition of the railroad tracks. Equipped with Real-Time Kinematics (RTK), the UAS provides high-precision geospatial referencing. During data collection, the UAS hovered along planned *S-shape* flight paths at controlled altitudes, capturing images of the railroad track from multiple angles. After scanning the railroad track using UAS, the collected images were imported into Agisoft Metashape Pro for 3D reconstruction. Using Structure from Motion (SfM) and Multi-View Stereo (MVS) algorithms, the software reconstructed a detailed 3D point cloud and generated a high-resolution model of the railroad, as shown in Figure 2(a). Then, CloudCompare was used to manually label the point cloud data. Operators carefully selected points corresponding to different structural components of the railroad, such as rails and crossties, and assigned labels to each category, as shown in Figure 2(b). In Figure 2(b), red points represent rails, while blue points represent crossties. Labeling a single railroad point cloud typically took around 15 minutes, depending on the complexity and density of the data. While manageable for a small number of samples, this manual process quickly becomes time-consuming and labor-intensive and may introduce manual labeling errors when scaling up for larger datasets. This highlights the need for more efficient labeling methods, such as incorporating synthetic BIM-generated point clouds to supplement the training data and reduce the reliance on manual annotation.

**BIM-Based Synthetic Data Generation and Automatic Labeling**

A railroad BIM model was created following standard railroad dimensions. Adopting the BIM model was primarily intended to realize the structural relationships among different components. Based on findings from previous studies, it was observed that point clouds generated from real-world data through photogrammetry tend to be surface models, while BIM-based models are often solid models [8]. To better match the characteristics of real-world scans and avoid inconsistencies in training, the BIM model was created as a surface structure originally in this paper. The constructed BIM surface model of a railroad track is shown in Figure 3(a). One of the benefits is that the BIM models can be easily duplicated in batches to expand the dataset size. Each component of the model can be quickly exported as an individual *.obj* file and then converted into a point cloud format using CloudCompare, without introducing extra noise. Unlike data collected using UAS from the actual structures, manual selection of points is not necessary — the exported model components are labeled automatically and accurately. Figure 3(b) illustrates the labeled synthetic model, where red points represent rails, and blue points represent crossties with high density. Compared with the 3D model created from the actual structural data, integrating BIM model significantly improves labeling efficiency, avoids classification ambiguity, and eliminates labeling errors.

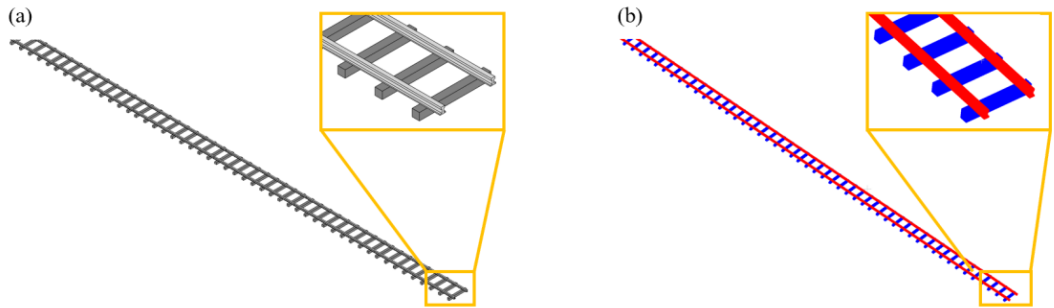

Figure 3. (a): Constructed surface BIM model of the railroad track with zoomed-in detail; (b): labeled synthetic point cloud with rails (red) and crossties (blue) with zoomed-in detail.

**Key Component Segmentation Based on Neural Network Architecture**

After obtaining real-world data collected by UAVs and synthetic data generated from BIM models, both datasets were adopted to train a segmentation model. Based on how the data is organized deep learning networks designed for semantic point cloud segmentation can be divided into two main categories: projection-based methods and point-based methods [9]. Projection-based approaches typically adopt well-established 2D networks to handle point cloud data. In contrast, point-based methods, such as the widely known PointNet [10], apply convolutional layers directly on individual points and their local neighborhoods to capture local features. Building on PointNet algorithm, many researchers have further developed global aggregation networks that progressively learn contextual information across the entire scene. Since then, PointNet has inspired many researchers to develop network models that progressively learn contextual information across entire scenes [11]. PointNet++ algorithm [12] is one such improvement, enhancing the original approach by capturing features at multiple scales. In this study, the architecture used for the task is based on PointNet++ architecture, which has been extensively tested and has demonstrated strong performance. It has been widely used across various industries for point cloud tasks including segmentation [13], [14].

TRAINING DATA PREPARATION

It was necessary to preprocess the point cloud data to ensure consistency across datasets before conducting data training. The real-world point clouds collected by UAVs contained tens of millions of points, while the synthetic BIM-generated point clouds reached hundreds of millions. This orders-of-magnitude difference in scale leads to a significant imbalance in data density, which can result in biased patch sampling and training inconsistency. Subsampling was introduced to reduce the point cloud density to address the data imbalance issue, bringing the real-world and synthetic datasets to a comparable scale. To further investigate whether a small-scale dataset could successfully train the model while saving training time, we prepared two levels of subsampled datasets. In the first set, each point cloud model contained between 2,000,000 and 5,000,000 points (the "million-level" dataset). While in the second set, each model contained between 200,000 and 500,000 points (the "hundred-thousand-level" dataset). Further, we applied random rotations along different axes to improve model robustness during data augmentation. Within each dataset, we created two subgroups: 1) point clouds were randomly rotated; 2) point clouds aligned with the railroad longitudinal direction. In addition, we further divided the datasets into groups that w/o BIM models to evaluate the contribution of BIM-generated synthetic data to model training. As a result, a total of 8 training groups were prepared for comparison, as summarized in Table I.

TABLE I. Training Dataset Configurations for Experimental Groups

| Group ID | Point Cloud Size | Rotation Strategy | BIM Data Included |
|---|---|---|---|
| G1 | Million-level | Random rotation | Yes |
| G2 | Million-level | Random rotation | No |
| G3 | Million-level | Railroad direction aligned | Yes |
| G4 | Million-level | Railroad direction aligned | No |
| G5 | Hundred-thousand-level | Random rotation | Yes |
| G6 | Hundred-thousand-level | Random rotation | No |
| G7 | Hundred-thousand-level | Railroad direction aligned | Yes |
| G8 | Hundred-thousand-level | Railroad direction aligned | No |

MODEL TRAINING AND EVALUATION

A patch-based point cloud sampling strategy was employed to prepare the training datasets. For the million-level datasets, patches containing 372,680 points were extracted, while patches of 37,268 points were used for the hundred-thousand-level datasets. This approach allowed efficient training at different density scales. PointNet++, a hierarchical deep learning architecture designed for point cloud processing, was used to learn multi-scale local features from the extracted patches and perform key structural component segmentation. The performance of the segmentation model was evaluated using real-world railroad point cloud data collected by the UAV, which was independent from the data used for training. The model's accuracy was assessed based on two commonly used metrics: Overall Accuracy (OA) and Intersection over Union (IoU) for each class. Overall Accuracy measures the proportion of correctly classified points across the entire dataset and is defined as:

$$OA = \frac{\text{Number of correctly predicted points}}{\text{Total number of points}} \quad (1)$$

Intersection over Union (IoU) was calculated separately for each class, representing the ratio between the number of correctly predicted points and the union of ground truth and predicted points for that class:

$$IoU_c = \frac{TP_c}{TP_c + FP_c + FN_c} \tag{2}$$

where, $TP_c$, $FP_c$, and $FN_c$ denote the true positives, false positives, and false negatives for class c, respectively. In addition, the Mean IoU (mIoU) was computed by averaging the IoUs across all classes:

$$mIoU = \frac{1}{C}\sum_{c=1}^{C} IoU_c \tag{3}$$

where $C$ is the total number of classes.

## RESULTS

To evaluate the effectiveness of the proposed training strategies and segmentation framework, we tested the trained models on a high-resolution point cloud (containing tens of millions of points) of a wood crossties railroad. This realistic and detailed dataset provided a reliable basis for assessing model performance. To compare the effectiveness of various training strategies, all eight dataset groups summarized in Table I were evaluated using three key indicators: training time, Overall Accuracy (OA), and mean Intersection over Union (mIoU). The results, presented in Table II, highlight how data size, rotation strategy, and the inclusion of BIM data influence segmentation performance.

As summarized in Table II, the inclusion of synthetic BIM point clouds (G1, G3, G5) in the training datasets led to consistently better performance in both OA and mIoU. These results suggest that BIM data enhances the model's ability to generalize to real-world scenarios. In addition, training with randomly rotated patches (G1, G5) yielded better results than training with direction-aligned data. The use of rotation augmentation enhanced the model's capability to capture structural variations, leading to improved segmentation performance. Comparing different dataset scales, models trained with million-level point clouds produced higher accuracy on the large-scale evaluation dataset. However, training with million-level data incurred significantly longer training time, exceeding 40 minutes per experiment (G1, G3). In contrast, models trained with hundred-thousand-level data (G5) reduced training time to a few minutes, although with a moderate drop in segmentation accuracy. Notably, the hundred-thousand-level dataset combined with BIM augmentation (G5) achieved an OA of 82.64% and an mIoU of 0.6920, demonstrating that synthetic data can help offset limited dataset size while greatly reducing training time. The corresponding segmentation results are shown in Figure 4(a), where green and red indicate the predicted rails and crossties, respectively. The model accurately identified key structural components, illustrating the effectiveness of combining BIM data and rotation augmentation for efficient large-scale segmentation. These results highlight the effectiveness of combining synthetic BIM data and rotation augmentation, particularly when balancing segmentation performance and computational efficiency in large-scale infrastructure applications.

TABLE II. Segmentation Performance Across Different Training Configurations

| Group ID | Data Size | Rotation Type | BIM Included | Training Time | OA | mIoU |
|---|---|---|---|---|---|---|
| G1 | Million-level | Random rotation | Yes | 43'27" | 85.94% | 0.7497 |
| G2 | Million-level | Random rotation | No | 24'17" | 76.92% | 0.6032 |
| G3 | Million-level | Direction aligned | Yes | 49'29" | 84.49% | 0.7308 |
| G4 | Million-level | Direction aligned | No | 22'22" | 69.43% | 0.4963 |
| G5 | Hundred-thousand-level | Random rotation | Yes | 3'40" | 82.64% | 0.6920 |
| G6 | Hundred-thousand-level | Random rotation | No | 2'58" | 66.73% | 0.4425 |
| G7 | Hundred-thousand-level | Direction aligned | Yes | 3'40" | 47.03% | 0.2522 |
| G8 | Hundred-thousand-level | Direction aligned | No | 2'55" | 50.09% | 0.2504 |

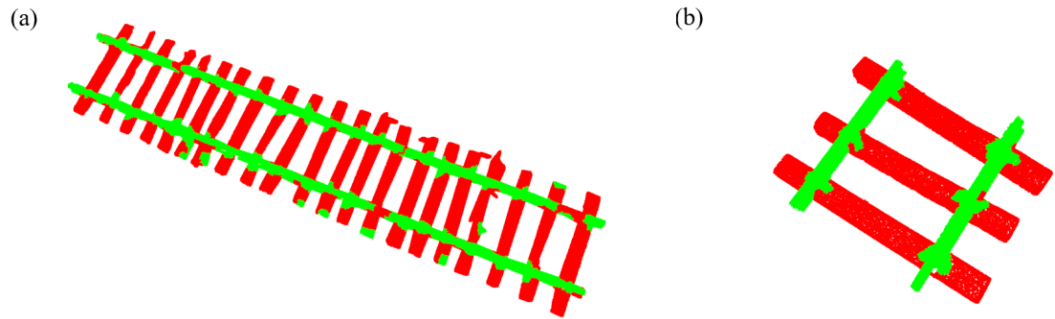

Figure 4. Segmentation results: (a) wood-tie railroad using hundred-thousand-level random rotation with BIM training; (b) concrete-tie railroad.

Furthermore, the trained model was evaluated on railroad tracks constructed with different crosstie materials, including concrete crossties, to assess its generalization performance. Representative segmentation results on concrete ties railroads sample are presented in Figure 4(b). The model also correctly segmented the concrete ties railroad section, with rails and crossties clearly distinguished. The result reflects the model's ability to capture structural features across different material types.

**CONCLUSION**

Efficient segmentation of key components from 3D point clouds is essential for railroad infrastructure health monitoring, however, manual annotation remains labor-intensive and inconsistent. To address this, we propose a method that integrates synthetic BIM-generated data with real-world UAV scans, significantly reducing labeling effort while maintaining high segmentation accuracy. The framework enables fast model convergence and performs well on high-resolution point clouds. Comparative experiments show that adding automatically labeled BIM data enhances accuracy even with limited manual labeled real-world data input, while directional randomness improves generalization. Notably, using smaller-scale (hundred-thousand-level) datasets with BIM augmentation achieves effective segmentation of large-scale point clouds, while reducing training time compared to larger datasets. Owing to the structural consistency of railways, the trained model generalizes well into different crosstie types, including wood and concrete, without retraining. Overall, the approach provides a scalable and practical solution for automated railroad infrastructure inspection.

## ACKNOWLEDGEMENT

The authors express their gratitude for the funding provided to support this study from USDOT - UTC, TRANS-IPIC and INDOT. The findings and opinions expressed in this article are those of the authors only and do not necessarily reflect the views of the sponsors.